\documentclass[10pt,conference]{IEEEtran}
\usepackage{cite}
\usepackage{amsmath,amssymb,amsfonts}
\usepackage{graphicx}
\usepackage{textcomp}
\usepackage{xcolor}
\usepackage{booktabs}
\usepackage{colortbl}
\usepackage{multirow}
\usepackage{hyperref}
\usepackage{orcidlink}

\graphicspath{{figures/}}

\newcommand{\graph}{\mathbf{G}}           
\newcommand{\grapht}{\mathbf{G}_t}        
\newcommand{\graphzero}{\mathbf{G}_0}     
\newcommand{\graphT}{\mathbf{G}_T}        
\newcommand{\nodem}{\mathbf{X}}           
\newcommand{\edget}{\mathbf{E}}           
\newcommand{\adj}{\mathbf{A}}             
\newcommand{\edgeset}{\mathcal{E}}        

\newcommand{\QtX}{\mathbf{Q}_t^X}         
\newcommand{\QtE}{\mathbf{Q}_t^E}         

\newcommand{\policy}{\pi_\theta}          
\newcommand{\reward}{r}                    
\newcommand{\rewardG}{r(\graphzero)}      
\newcommand{\advantage}{\hat{A}}          
\newcommand{\state}{\mathbf{s}}           
\newcommand{\action}{\mathbf{a}}          
\newcommand{\traj}{\boldsymbol{\tau}}     

\newcommand{\lossDGPO}{\mathcal{L}_{\text{DGPO}}}  
\newcommand{\lossCE}{\mathcal{L}_{\text{CE}}}      

\newcommand{\penc}{\mathbf{p}}            
\newcommand{\topo}{\sigma}                
\newcommand{\threshold}{\mathcal{T}}      

\newcommand{\E}{\mathbb{E}}              
\newcommand{\R}{\mathbb{R}}              

\begin{document}

\title{DGPO: RL-Steered Graph Diffusion for\\Neural Architecture Generation}

\author{\IEEEauthorblockN{Aleksei Liuliakov\,\orcidlink{0000-0003-4676-9272},
Luca Hermes\,\orcidlink{0000-0002-7568-7981},
Barbara Hammer\,\orcidlink{0000-0002-0935-5591}}
\IEEEauthorblockA{Machine Learning Group, Bielefeld University, Germany\\
\{aliuliakov\textbar lhermes\textbar bhammer\}@techfak.uni-bielefeld.de}}

\maketitle

\begin{quote}
\small\textit{Published at IEEE IJCNN 2026 (WCCI). \textcopyright~2026 IEEE.}
\end{quote}

\begin{abstract}
Reinforcement learning fine-tuning has proven effective for steering generative diffusion models toward desired properties in image and molecular domains.
Graph diffusion models have similarly been applied to combinatorial structure generation, including neural architecture search (NAS).
However, neural architectures are directed acyclic graphs (DAGs) where edge direction encodes functional semantics such as data flow - information that existing graph diffusion methods, designed for undirected structures, discard.
We propose Directed Graph Policy Optimization (DGPO), which extends reinforcement learning fine-tuning of discrete graph diffusion models to DAGs via topological node ordering and positional encoding.
Validated on NAS-Bench-101 and NAS-Bench-201, DGPO matches the benchmark optimum on all three NAS-Bench-201 tasks (91.61\%, 73.49\%, 46.77\%).
The central finding is that the model learns transferable structural priors: pretrained on only 7\% of the search space, it generates near-oracle architectures after fine-tuning, within 0.32 percentage points of the full-data model and extrapolating 7.3 percentage points beyond its training ceiling.
Bidirectional control experiments confirm genuine reward-driven steering, with inverse optimization reaching near random-chance accuracy (9.5\%).
These results demonstrate that reinforcement learning-steered discrete diffusion, once extended to handle directionality, provides a controllable generative framework for directed combinatorial structures.
\end{abstract}


\section{Introduction}
\label{sec:introduction}

Diffusion models have emerged as powerful generative frameworks for learning complex distributions, with recent extensions to graph-structured data~\cite{Vignac2022DiGressDD} enabling direct generation of molecular graphs and other combinatorial objects.
Reinforcement learning (RL) fine-tuning steers pretrained diffusion models toward desired properties: Denoising Diffusion Policy Optimization (DDPO)~\cite{Black2024_DDPO} introduced this for image generation, and Graph Diffusion Policy Optimization (GDPO)~\cite{Liu2025_GDPO} extended it to molecular graphs.
In the graph domain, however, these methods have been limited to undirected structures.

Many important domains involve directed acyclic graphs (DAGs): neural architectures encode data flow through directed edges, causal networks represent asymmetric dependencies, and computational workflows specify ordered execution.
In such graphs, edge direction encodes functional semantics that is destroyed by symmetric treatment.
Neural architecture search (NAS)~\cite{Zoph2016NeuralAS, White2023} - the problem of automatically discovering high-performing neural network architectures from a structured search space -exemplifies this challenge: existing generative approaches either condition generation on target accuracy~\cite{Asthana2024} or build on graph diffusion models designed for undirected structures~\cite{Vignac2022DiGressDD}, forgoing directional information.
Extending RL-steered discrete diffusion to DAGs requires respecting and reconstructing this directional structure.

We propose Directed Graph Policy Optimization (DGPO), extending GDPO~\cite{Liu2025_GDPO}, a graph diffusion RL fine-tuning framework, to directed acyclic graphs via topological node ordering and positional encoding.
This enables the discrete diffusion process to respect directionality while preserving the reward-driven steering mechanism of the underlying framework.

We validate DGPO on two public NAS benchmarks, NAS-Bench-101~\cite{Ying2019_NASBench101} and NAS-Bench-201~\cite{Dong2020NASBench201ET}, achieving competitive architecture generation that reaches the NB201 benchmark optimum on all three tasks (91.61\%, 73.49\%, and 46.77\%).
We find that the model can acquire transferable structural priors: pretrained on only 7\% of the search space, it generates near-oracle architectures after RL fine-tuning, within 0.32 percentage points of the full-data model and extrapolating up to 7.3pp beyond its training ceiling.
We provide additional evidence for reward-driven steering via bidirectional control, where inverting the reward signal drives generation toward near random-chance accuracy (9.5\%), and we also show how the framework extends to multi-objective reward formulations.

Our contributions are:
\begin{enumerate}
\item \textbf{Methodological:} DGPO extends GDPO to directed acyclic graphs via topological node ordering and positional encoding, enabling controllable generation of DAG-structured objects.
\item \textbf{Empirical (primary):} We show evidence that the resulting model learns transferable structural priors: a model trained on only 7\% of the NAS search space generates near-oracle architectures after RL fine-tuning (within 0.32pp), with up to $+$7.3pp extrapolation beyond its training ceiling.
\item \textbf{Empirical (supporting):} We provide additional evidence for the steering mechanism through bidirectional control (inverse reaches near random chance at 9.5\%) and show extension to multi-objective reward formulations.
\end{enumerate}

Code and pretrained models are available at:\par
\url{https://github.com/AlekseiLiu/DGPO}.

\section{Related Work}
\label{sec:related_work}

\textbf{Graph diffusion models.}
Diffusion models have been extended to graph-structured data through both discrete and continuous formulations~\cite{Fan2023GenerativeDM}.
DiGress~\cite{Vignac2022DiGressDD} introduced discrete denoising diffusion for graphs, applying categorical noise transitions to node and edge attributes with a graph transformer denoiser.
Autoregressive variants~\cite{Kong2023AutoregressiveDM} generate graphs sequentially but sacrifice parallelism.
DGPO builds on DiGress as its discrete diffusion backbone.

\textbf{RL fine-tuning of diffusion models.}
Reinforcement learning can steer pretrained diffusion models toward desired properties by framing denoising as a Markov decision process~\cite{Uehara2024_RLFTTutorial}.
DDPO~\cite{Black2024_DDPO} introduced this approach for image generation, optimizing human preference scores via policy gradients.
DPOK~\cite{Fan2024_DPOK} adds KL regularization for text-to-image alignment.
GDPO~\cite{Liu2025_GDPO} adapts the framework to molecular graph generation using reward-weighted denoising trajectories.
These methods operate on undirected structures (images, molecules); DGPO extends the GDPO framework to directed acyclic graphs.

\textbf{Neural architecture search.}
NAS has been addressed through RL controllers~\cite{Zoph2016NeuralAS}, evolutionary algorithms~\cite{Real2019_REA}, differentiable relaxations such as DARTS~\cite{Liu2018DARTSDA}, and predictor-based methods including BANANAS~\cite{White2021_BANANAS}.
Generative approaches learn to directly produce architectures: D-VAE~\cite{Zhang2019DVAEAV} uses variational autoencoders for directed graph generation, DiffusionNAG~\cite{An2023DiffusionNAGTN} applies diffusion models for task-guided generation, AG-Net~\cite{Lukasik2022_AGNet} uses autoregressive generation with a learned surrogate, and GraphPNAS~\cite{Li2022} learns distributions over high-performing architectures.
DiNAS~\cite{Asthana2024} conditions a graph diffusion model on target accuracy to generate architectures directly.
DGPO differs from conditioning-based approaches by steering the generation distribution via RL fine-tuning, enabling capabilities beyond accuracy targeting: bidirectional control, out-of-distribution discovery from filtered data, and multi-objective optimization.

We extend an existing RL fine-tuning framework (GDPO) to a structurally different graph class (DAGs), rather than proposing a new RL algorithm or diffusion architecture.
The contribution lies in enabling RL-steered discrete diffusion for directed combinatorial structures and providing empirical evidence for transferable structural priors in the resulting generative model.

\section{Method}
\label{sec:method}

We first review the discrete diffusion and RL fine-tuning foundations that DGPO builds on, then present our extension to directed acyclic graphs.

\subsection{Preliminaries}
\label{sec:preliminaries}

\textbf{Discrete graph diffusion.}
DiGress~\cite{Vignac2022DiGressDD} defines a discrete diffusion process over graphs $\graph = (\nodem, \edget)$ with $n$ nodes.
Node attributes $\nodem \in \{1,\ldots,a\}^n$ take one of $a$ categorical types; edge attributes $\edget \in \{0,1,\ldots,b\}^{n \times n}$ take one of $b$ types plus an \emph{absent} state.
Attributes are categorical because the diffusion process operates via discrete Markov transition matrices over a finite state set at each step, rather than adding continuous noise.
The forward process independently corrupts each attribute of the clean graph $\graphzero$:
\begin{equation}
q(\grapht \mid \graph_{t-1}) = \!\prod_{i} q(x_i^t \mid x_i^{t-1})\!\prod_{i<j} q(e_{ij}^t \mid e_{ij}^{t-1}),
\label{eq:forward}
\end{equation}
where the edge product runs over unique pairs $i < j$ since the diffusion operates on a symmetric (undirected) intermediate representation.
Each factor is a categorical draw: $x_i^t \sim \mathrm{Cat}\!\bigl((\mathbf{x}_i^{t-1})^\top \QtX\bigr)$ and $e_{ij}^t \sim \mathrm{Cat}\!\bigl((\mathbf{e}_{ij}^{t-1})^\top \QtE\bigr)$, with $\QtX \in \R^{a \times a}$ and $\QtE \in \R^{(b{+}1) \times (b{+}1)}$ the per-step transition matrices for node and edge attributes, respectively.
Over $T$ steps this produces $\graphzero, \graph_1, \ldots, \graphT$, with $\graphT$ approaching the limit distribution of the transition model (uniform or dataset marginals; following DiGress, we use marginals in our implementation~\cite{Vignac2022DiGressDD}).
A denoising network $\varphi_\theta$, parameterized by $\theta$, learns to invert this corruption by predicting $\graphzero$ from the noisy input $\grapht$ at each step~\cite{Austin2021D3PM}.
Graphs are generated by sampling $\graphT$ from the prior and iteratively applying the learned reverse transitions.

\textbf{Reinforcement learning fine-tuning of diffusion models.}
GDPO~\cite{Liu2025_GDPO} formulates the denoising process as a $T$-step Markov decision process: the state is $\state_t = (\graph_{T-t},\, T{-}t)$, the action is $\action_t = \graph_{T-t-1}$, the policy $\policy$ is the denoising model, and a sparse reward $\rewardG$ is assigned only to the final generated graph.
The objective is to maximize expected reward $\E_{\traj \sim \policy}[\rewardG]$ over denoising trajectories $\traj = (\graphT, \ldots, \graphzero)$.
DDPO~\cite{Black2024_DDPO} introduced this MDP framing for image diffusion; GDPO extends it to molecular graphs using an \emph{eager policy gradient} that directly optimizes $\nabla_\theta \log p_\theta(\graphzero \mid \grapht)$ at each timestep, reducing gradient variance compared to the standard REINFORCE estimator~\cite{Liu2025_GDPO}.

\subsection{DGPO: Extending GDPO to Directed Acyclic Graphs}
\label{sec:method_dgpo}

GDPO was developed for undirected molecular graphs where edge semantics are symmetric.
In directed acyclic graphs (DAGs) - neural architectures, causal networks, computational workflows - edge direction encodes functional semantics such as data flow or causation.
Treating directed edges as symmetric destroys this information.
We propose Directed Graph Policy Optimization (DGPO), which extends GDPO to DAGs via three components.

\textbf{Topological node ordering.}
Before diffusion, we apply a topological ordering $\topo$ to the DAG nodes such that $\forall (u,v) \in \edgeset{:}\; \topo(u) < \topo(v)$, where $\edgeset$ denotes the edge set of the DAG.
Under this ordering the adjacency matrix $\adj$ becomes strictly upper-triangular: $\adj_{ij} \neq 0 \Rightarrow i < j$, so all directional information is encoded in node indices.
NAS benchmarks store architectures in a consistent DAG order (input node first, output node last); for general DAGs, any standard topological sort (e.g., Kahn's algorithm) can be applied in preprocessing.

\textbf{Positional encoding.}
We augment each node's feature vector with a sinusoidal positional encoding $\penc_i \in \R^d$ derived from its topological position $i$~\cite{Dwivedi2020_GraphTransformer}:
\begin{equation}
\penc_{i,2k} = \sin\!\bigl(i / 10000^{2k/d}\bigr),\quad
\penc_{i,2k+1} = \cos\!\bigl(i / 10000^{2k/d}\bigr),
\label{eq:pos_enc}
\end{equation}
for $k = 0, \ldots, d/2{-}1$, where $d$ is the hidden dimension of the graph transformer.
The encoding is added to node features after the input projection and before the first transformer layer: $\mathbf{h}_i \leftarrow \mathbf{h}_i + \penc_i$.
This provides $\varphi_\theta$ with explicit positional information, enabling it to distinguish node roles (e.g., early- vs.\ late-layer operations) and reconstruct directed edges from the noisy intermediate graph.

\textbf{DAG recovery.}
During diffusion, intermediate graphs are treated as undirected to remain compatible with the DiGress backbone, which symmetrizes edge noise.
After the final denoising step produces a full adjacency matrix $\adj \in \{0,1,\ldots,b\}^{n \times n}$, we recover a valid DAG by retaining only upper-triangular entries: $\adj^\text{dag}_{ij} = \adj_{ij}$ if $i < j$, and $0$ otherwise (diagonal masked, as self-loops are invalid in the NAS benchmarks).
This projection guarantees acyclicity by construction and requires no learned components.

Together, topological node ordering and positional encoding enable the DiGress backbone and GDPO fine-tuning framework - designed for undirected graphs - to handle DAGs without modifying the diffusion or RL architectures.
The extension is domain-agnostic: any DAG-structured combinatorial problem can be addressed by providing an appropriate reward signal.

\subsection{Training Objective}
\label{sec:method_objective}

Following GDPO~\cite{Liu2025_GDPO}, the DGPO training objective is a reward-weighted cross-entropy loss over denoising trajectories with advantage normalization:
\begin{equation}
\lossDGPO(\theta) = \frac{1}{K} \sum_{k=1}^{K} \frac{T}{|\mathcal{T}_k|} \sum_{t \in \mathcal{T}_k} \advantage_k \cdot \lossCE\!\bigl(\varphi_\theta(\graph_t^{(k)}),\, \graphzero^{(k)}\bigr),
\label{eq:dgpo_loss}
\end{equation}
where $K$ is the number of trajectories per batch, $\mathcal{T}_k$ is a uniformly sampled subset of timesteps for trajectory $k$, and the cross-entropy decomposes over nodes and edges:
\begin{equation}
\lossCE = \sum_{i} \text{CE}(x_i,\, \hat{x}_i) + \lambda \sum_{i,j} \text{CE}(e_{ij},\, \hat{e}_{ij}).
\end{equation}
The advantage estimate normalizes the reward signal:
\begin{equation}
\advantage_k = \text{clip}\!\left(\frac{\reward_k - \bar{\reward}}{\sigma_\reward},\, {-}5,\, 5\right),
\end{equation}
where $\bar{r}$ and $\sigma_r$ are the running mean and standard deviation of the reward.
The cross-entropy provides per-timestep $\nabla_\theta \log p_\theta(\graphzero \mid \grapht)$ gradients, while the advantage reweights each trajectory by its relative quality, yielding a REINFORCE-style policy gradient with variance reduction.

The reward signal admits several formulations: single-objective ($\rewardG$ = normalized validation accuracy), inverse ($-\rewardG$ for mechanism validation), and multi-objective ($\sum_i w_i \cdot r_i(\graphzero)$ for compound tasks).

\subsection{Two-Phase Training}
\label{sec:two_phase}

\textbf{Phase~1: Pretraining.}
The diffusion model is trained on the benchmark architecture distribution using the denoising cross-entropy loss $\lossCE$, which trains $\varphi_\theta$ to predict $\graphzero$ from noisy inputs $\grapht$ at uniformly sampled timesteps $t \in \{1, \ldots, T\}$.
No reward signal or advantage term is used; the model learns the structural grammar of the search space - valid node types, connectivity patterns, and size distributions - purely from the data distribution.

\textbf{Phase~2: RL fine-tuning.}
Starting from the pretrained checkpoint, the model is fine-tuned using the DGPO objective~\eqref{eq:dgpo_loss}.
Layer freezing (75\% of parameters) stabilizes training and prevents catastrophic forgetting; gradient accumulation compensates for the small batch sizes required by online reward evaluation.
This phase shifts the generation distribution toward the reward signal while preserving the structural validity learned in Phase~1.

\section{Experiments}
\label{sec:experiments}

\subsection{Experimental Setup}
\label{sec:setup}

\textbf{Benchmarks.}
We evaluate on NAS-Bench-101 (NB101)~\cite{Ying2019_NASBench101} and NAS-Bench-201 (NB201)~\cite{Dong2020NASBench201ET}.
NB101 contains 423,624 unique architectures represented as directed acyclic graphs with up to seven nodes and 21 possible edges, evaluated on CIFAR-10.
NB201 defines a cell-based search space of 15,625 architectures evaluated on three tasks: CIFAR-10, CIFAR-100, and ImageNet-16-120.

\textbf{Model architecture.}
DGPO uses a DiGress backbone~\cite{Vignac2022DiGressDD} with eight Transformer layers, $T{=}800$ diffusion steps, and a cosine noise schedule.
We augment each node with a topological positional encoding that captures its position in the topological ordering (Section~\ref{sec:method_dgpo}).
At generation time, valid DAGs are recovered via upper-triangular projection of the generated adjacency matrix.

\textbf{Training protocol.}
Training follows a two-phase procedure.
Phase~1 pretrains the diffusion model on the full benchmark architecture distribution using cross-entropy loss.
Phase~2 applies RL fine-tuning (RL-FT) for 60 epochs with batch size $K{=}15$, learning rates of $7{\times}10^{-7}$ (NB101) and $5{\times}10^{-7}$ (NB201), 75\% layer freezing, and AdamW optimization.
The training objective is a REINFORCE-style policy gradient with advantage normalization (Section~\ref{sec:method_objective}).

\textbf{Evaluation protocol.}
All experiments use three seeds (42, 123, 456); results are reported as mean~$\pm$~std.
Each evaluation draws 300 architectures from the current model.
NB101 and NB201 results are obtained via benchmark API lookup.

\textbf{Baselines.}
We compare against two internal baselines: \emph{random search} (uniform sampling from the search space) and \emph{pretrained-only} sampling (generation from the pretrained diffusion model without RL-FT).
External comparisons with prior NAS methods are presented in Section~\ref{sec:baseline_results}.


\subsection{Baseline Results}
\label{sec:baseline_results}

\begin{figure*}[t]
\centering
\includegraphics[width=\textwidth]{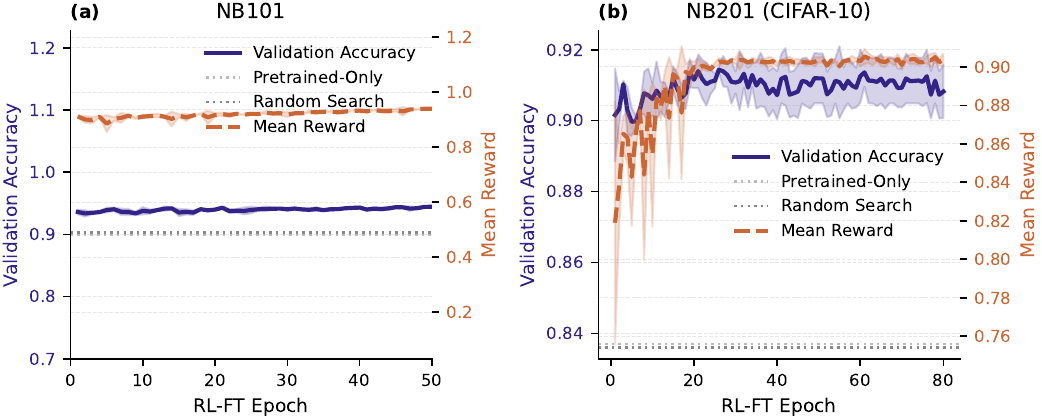}
\caption{Training dynamics of DGPO on (a) NB101 and (b) NB201 (CIFAR-10): validation accuracy (solid, left axis) and mean reward (dashed, right axis) over RL-FT epochs, with ${\pm}1\sigma$ bands across 3 seeds. Horizontal lines: random search and pretrained-only baselines. Both metrics converge reliably, confirming that RL fine-tuning steers the generation distribution toward higher-quality architectures.}
\label{fig:training_dynamics}
\end{figure*}

DGPO steers DAG generation toward high-accuracy architectures on both NB101 and NB201.
Fig.~\ref{fig:training_dynamics} shows training dynamics over RL-FT epochs: validation accuracy increases over training on both benchmarks while mean reward rises in tandem, with narrow ${\pm}1\sigma$ bands across three seeds suggesting stable convergence.

\begin{figure*}[t]
\centering
\includegraphics[width=\textwidth]{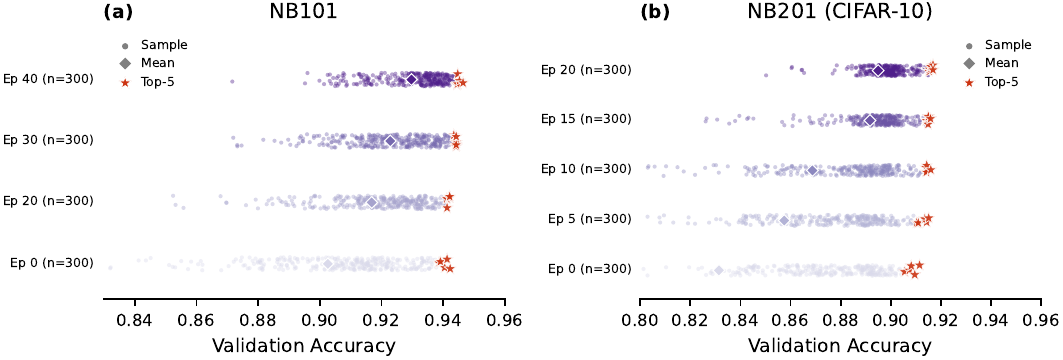}
\caption{Distribution of generated architectures over RL-FT epochs on (a) NB101 and (b) NB201 (CIFAR-10). Each strip shows 300 sampled architectures (dots) at a given epoch, with mean accuracy (diamond) and top-5 architectures (stars). As training progresses (bottom to top), the mean and overall sample density shift markedly toward higher accuracy, demonstrating that RL fine-tuning reshapes the generative distribution rather than merely selecting isolated high performers. Single seed (42); $n{=}300$ samples per epoch.}
\label{fig:distribution_shift}
\end{figure*}

Fig.~\ref{fig:distribution_shift} illustrates the underlying mechanism: RL-FT progressively shifts the generated architecture distribution toward higher accuracy, concentrating probability mass in the high-performing region.

Table~\ref{tab:compact_sota} compares DGPO against published NAS methods.
On NB201, DGPO matches the benchmark optimum on all three datasets: 91.61\% (CIFAR-10), 73.49\% (CIFAR-100), and 46.77\% (ImageNet-16-120), surpassing DiNAS~\cite{Asthana2024} on ImageNet-16-120 (46.77\% vs 46.66\%).
On NB101, DGPO achieves 94.50\%, competitive with established methods but below DiNAS (94.98\%).
On NB101, several established methods achieve similar accuracy levels within a narrow range; DGPO's distinctive value on this benchmark lies in the controllable distribution steering demonstrated in Sections~\ref{sec:ood_discovery} and~\ref{sec:steering_versatility}.

\begin{table}[t]
\caption{Comparison with SOTA NAS methods. Max validation accuracy (\%)
on NB101 and NB201. Prior results from
\cite{Asthana2024} (mean over 10 runs); DGPO: mean$\pm$std over 3 seeds
(NB101) / deterministic API lookup (NB201). * = benchmark optimum.
DGPO uses ${\sim}$2k queries; other methods use 150--192.}
\centering
\small
\begin{tabular}{l c c c c}
\hline
\multirow{2}{*}{Method} & NB101 & \multicolumn{3}{c}{NB201 Val (\%)} \\
                        & Val   & C-10 & C-100 & IN-16 \\
\hline
Optimum         & 95.06 & 91.61 & 73.49 & 46.77 \\
\hline
Random Search   & 94.31 & 91.12 & 72.08 & 45.97 \\
Reg.\ Evol.     & 94.47 & --    & --    & --    \\
Bayesian Opt.   & 94.57 & 91.54 & 73.26 & 46.43 \\
BANANAS         & 94.73 & 91.56 & 73.49*& 46.65 \\
AG-Net          & 94.90 & 91.60 & 73.49*& 46.64 \\
DiNAS           & \textbf{94.98} & 91.61*& 73.49*& 46.66 \\
\hline
\textbf{DGPO}
  & $94.50{\scriptstyle\pm0.02}$
  & \textbf{91.61*}
  & \textbf{73.49*}
  & \textbf{46.77*}
  \\
\hline
\end{tabular}
\label{tab:compact_sota}
\end{table}

The additional query cost (${\sim}$2k vs 150--192) reflects the online RL paradigm: rather than a fixed-budget search, DGPO fine-tunes the generative distribution itself.
This distribution-level steering enables the capabilities demonstrated in the following sections - bidirectional control, out-of-distribution discovery, and multi-objective optimization - which are not available to conditioning-based methods.


\subsection{Transferable Structural Priors}
\label{sec:ood_discovery}

Does the generative model learn compositional structural priors, or does it merely memorize the training distribution?
To answer this, we filter the pretraining data to architectures below a quality threshold $\threshold$, removing all high-accuracy instances, and then apply RL-FT to the filtered model.
Specifically, we retain only architectures with validation accuracy below $\threshold{=}0.87$ on NB101 (7\% of data; ${\sim}$30k of 423k) and below $\threshold{=}0.85$ on NB201 (29\%; ${\sim}$4,500 of 15,625).
Out-of-distribution (OOD) architectures - those with validation accuracy above $\threshold$ - are entirely absent from the filtered pretraining set.

\textbf{Filtered vs full pretraining.}
Table~\ref{tab:filtered_vs_full} compares RL-FT results starting from the filtered versus full pretrained model.
On NB101, the filtered model achieves 94.18\%, only 0.32 percentage points below the full model's 94.50\%.
On NB201, the filtered model reaches 91.71\%, within noise of the full model (91.70\%).
Removing 93\% of training data costs only 0.32pp in generation quality on NB101 and is negligible on NB201, indicating that weak-region data suffices for acquiring structural priors.

\begin{table}[t]
\caption{RL-FT on full vs filtered pretraining: max validation accuracy (mean $\pm$ std, 3 seeds).
Filtered models are pretrained exclusively on architectures below threshold $\threshold$ (NB101: $\threshold{=}0.87$, 7\% of data; NB201: $\threshold{=}0.85$, 29\%).
Filtered pretraining loses only 0.32pp on NB101 and is within noise on NB201.}
\centering
\begin{tabular}{lcc}
\toprule
Method & NB101 & NB201 (C-10) \\
\midrule
Full Pretrain & \textbf{94.50 $\pm$ 0.02} & 91.70 $\pm$ 0.02 \\
Filtered Pretrain & 94.18 $\pm$ 0.33 & \textbf{91.71 $\pm$ 0.02} \\
\bottomrule
\end{tabular}
\label{tab:filtered_vs_full}
\end{table}

\begin{figure*}[t]
\centering
\includegraphics[width=\textwidth]{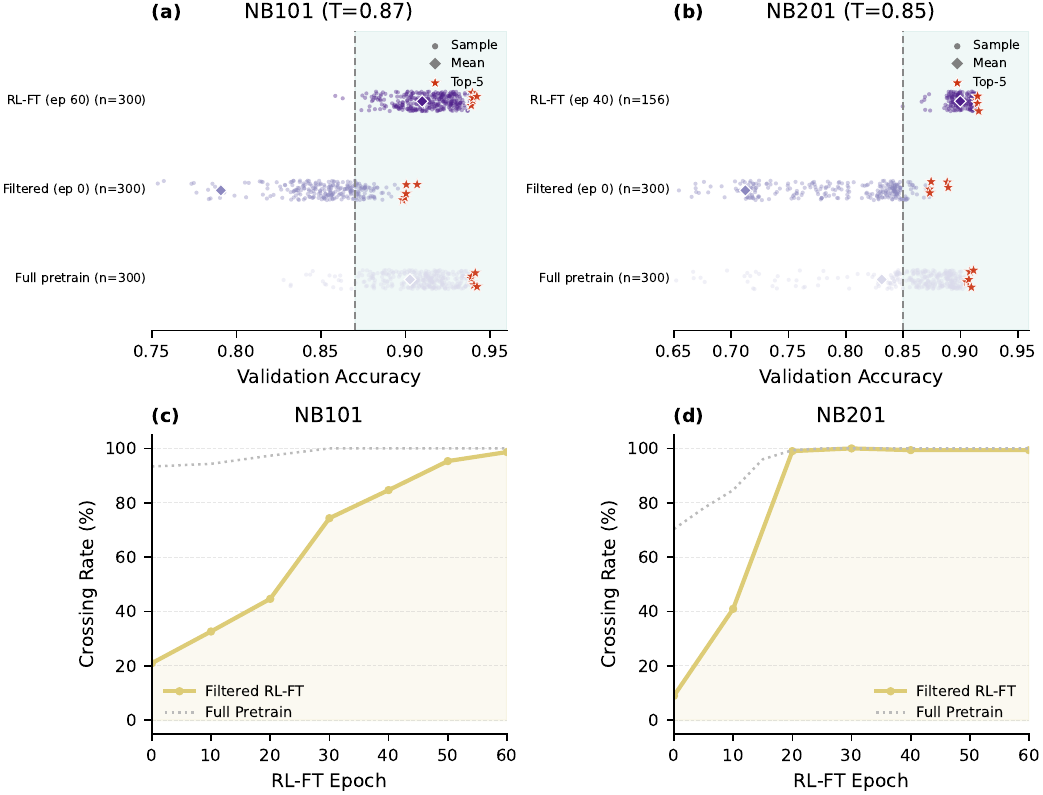}
\caption{OOD architecture discovery on (a,\,c) NB101 ($\threshold{=}0.87$) and (b,\,d) NB201 ($\threshold{=}0.85$).
Top: distribution comparison - full pretrain (reference), filtered pretrain (epoch~0), and RL-FT (final epoch). Shaded region marks OOD architectures above $\threshold$.
Bottom: threshold crossing rate over RL-FT epochs.
After pretraining on only sub-threshold architectures, RL-FT recovers above-threshold generation, demonstrating transferable structural priors.
Single seed (42) for distributions; crossing rates are 3-seed aggregates.}
\label{fig:ood_discovery}
\end{figure*}

\textbf{OOD architecture discovery.}
The filtered model generates architectures far above its training ceiling.
Fig.~\ref{fig:ood_discovery} (top row) visualizes this: the filtered pretrained distribution (middle strip) clusters below $\threshold$, but after RL-FT (top strip) the distribution recovers to resemble the full pretrained reference (bottom strip) - despite never having seen above-$\threshold$ architectures during pretraining.
The bottom row tracks the threshold crossing rate (percentage of generated samples with accuracy ${\geq}\threshold$) over RL-FT epochs.
On NB101, the filtered model starts at ${\sim}$21\% above $\threshold$ and climbs to ${\sim}$99\% after RL-FT.
On NB201, it starts at ${\sim}$9\% and reaches ${\sim}$99\%.

\textbf{Per-seed consistency.}
Across all three seeds, the OOD discovery rate is 100\% on both benchmarks: every seed generates architectures above $\threshold$ after RL-FT.
On NB101, the best generated architecture reaches $94.27 \pm 0.11\%$, extrapolating $+$7.27pp above the training ceiling, with a mean crossing rate of $67.7 \pm 3.2\%$.
On NB201, the best architecture reaches $91.68 \pm 0.06\%$, matching the benchmark oracle, with crossing rate $79.2 \pm 1.7\%$ and an OOD lift of $0.77 \pm 0.02$ (OOD architectures are 77\% more likely to appear among the top-performing generated samples).

\textbf{Interpretation.}
These results provide evidence for compositional generalization in the diffusion model: pretraining on weak architectures captures local structural motifs - connectivity patterns, operation combinations - that, when recombined under RL-FT guidance, produce high-quality architectures absent from the training set.
The model extrapolates beyond its training support by recombining learned primitives, not by memorizing successful configurations.
We use filtering as a controlled stress test; the threshold $\threshold$ is fixed before RL-FT and not tuned post hoc.


\subsection{Steering Versatility}
\label{sec:steering_versatility}

\begin{figure*}[t]
\centering
\includegraphics[width=\textwidth]{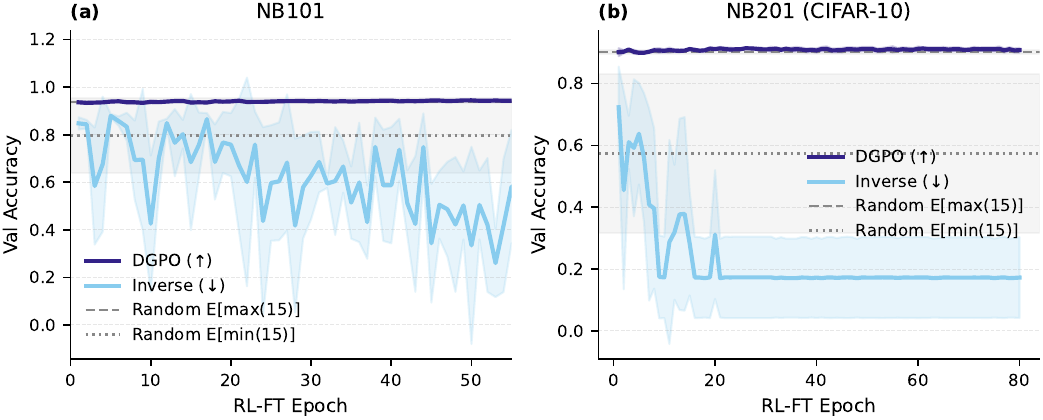}
\caption{Bidirectional steering on (a) NB101 and (b) NB201 (CIFAR-10): forward (maximize, $\uparrow$) and inverse (minimize, $\downarrow$) DGPO trajectories over RL-FT epochs, with ${\pm}1\sigma$ bands (3 seeds).
Dashed/dotted lines: expected max/min of random batches ($N{=}15$, bootstrap $K{=}10{,}000$).
The inverse trajectory converges to near random-chance accuracy (${\sim}$9.5\%), supporting reward-driven distribution steering.}
\label{fig:bidirectional}
\end{figure*}

\textbf{Bidirectional control.}
The same RL-FT framework steers generation in either direction, validating that the reward signal drives genuine structure--performance learning rather than exploitation of the pretraining distribution.
Fig.~\ref{fig:bidirectional} shows forward (maximize accuracy) and inverse (minimize accuracy) DGPO trajectories.
The forward trajectory rises above the expected best-of-batch under random sampling ($\E[\max(15)]$, computed via bootstrap with $K{=}10{,}000$ resamples).
Negating the reward signal causes the model to generate progressively worse architectures: the inverse trajectory converges toward near random-chance accuracy (9.51\% on NB101, 9.71\% on NB201; chance is 10\% for 10-class classification).
This provides evidence against explanations where the model merely exploits the pretraining distribution: with a negated reward, the same model is driven toward architectures it was not encouraged to generate during pretraining.

On NB201, one of three seeds temporarily lingers near a secondary basin (${\sim}$59\%), while others reach the floor, suggesting bimodal structure in the inverse optimization landscape.
All seeds reach the global minimum during training; the wider error band reflects basin transitions rather than method failure.

\textbf{Multi-objective steering.}
We extend DGPO to multi-objective (MO) optimization on NB201 by using a compound reward: the weighted sum of normalized validation accuracies on CIFAR-10, CIFAR-100, and ImageNet-16-120.
MO-DGPO improves all three tasks simultaneously, with the composite reward increasing by 26\%.
The Pareto hypervolume reaches 0.9901, matching the union of three separate single-task runs.
MO-DGPO achieves comparable front quality in a single joint run, demonstrating that the RL-FT framework generalizes naturally from single-task to multi-objective formulations without architectural changes.

An adversarial weight configuration (negative weight on CIFAR-10) reveals strong task correlation in the NB201 search space: penalizing one task simultaneously degrades others.
This confirms the known correlation structure in NB201 and illustrates that DGPO can probe task relationships via reward engineering.

\section{Conclusion}
\label{sec:conclusion}

We presented DGPO, extending RL-steered discrete graph diffusion to directed acyclic graphs via topological node ordering and positional encoding.
Validated on NAS-Bench-101 and NAS-Bench-201, DGPO achieves competitive architecture generation and provides evidence of transferable structural priors: a model pretrained on only weak-region data can generate near-oracle architectures after RL fine-tuning.
Bidirectional control and multi-objective steering further support the view that the reward signal drives structure--performance learning.

The approach incurs higher query cost than conditioning-based methods, reflecting the online RL paradigm, and has been validated on NAS benchmarks only; generalization to other DAG domains remains to be demonstrated.

Promising directions include reducing the query budget through offline RL or sample reuse strategies, scaling to open-ended search spaces such as NAS-Bench-301, and applying DGPO to other directed combinatorial domains including circuit design and causal discovery.

\bibliographystyle{IEEEtran}
\bibliography{references}

\end{document}